\documentclass[10pt,twocolumn,letterpaper]{article}

\usepackage[pagenumbers]{cvpr} 
\usepackage[dvipsnames]{xcolor}
\newcommand{\red}[1]{{\color{red}#1}}
\newcommand{\blue}[1]{{\color{blue}#1}}

\usepackage{multirow} 

\usepackage{verbatim}

\definecolor{cvprblue}{rgb}{0.21,0.49,0.74}
\usepackage[pagebackref,breaklinks,colorlinks,citecolor=cvprblue]{hyperref}
\usepackage{stmaryrd}
\usepackage{amsmath}
\usepackage{ulem}
\usepackage{wrapfig}

\def\paperID{3353} 
\def\confName{CVPR}
\def\confYear{2024}

\usepackage{amsmath,amsfonts,bm}

\def\eqref#1{equation~\ref{#1}}

\def\1{\bm{1}}

\DeclareMathAlphabet{\mathsfit}{\encodingdefault}{\sfdefault}{m}{sl}
\SetMathAlphabet{\mathsfit}{bold}{\encodingdefault}{\sfdefault}{bx}{n}

\title{Understanding the Multi-modal Prompts of the Pre-trained Vision-Language Model}
\author{
    Shuailei Ma \textsuperscript{\rm 1} ~
    Chen-Wei Xie\textsuperscript{\rm 2} ~ 
    Ying Wei\textsuperscript{\rm 1}\thanks{Corresponding author.  Work done when working as an intern in Alibaba Group.} ~ 
    Siyang Sun\textsuperscript{\rm 2}\\
    Jiaqi Fan\textsuperscript{\rm 1} ~
    Xiaoyi Bao\textsuperscript{\rm 3} ~
    Yuxin Guo\textsuperscript{\rm 3} 
    Yun Zheng\textsuperscript{\rm 2} \\
    \textsuperscript{\rm 1}\small{Northeast University, China}\quad \quad
    \textsuperscript{\rm 2}\small{Alibaba Group}\quad \quad
    \textsuperscript{\rm 3}\small{CASIA}\\
}

\begin{document}
\maketitle
\begin{abstract}
Prompt learning has emerged as an efficient alternative for fine-tuning foundational models, such as CLIP, for various downstream tasks. However, there is no work that provides a comprehensive explanation for the working mechanism of the multi-modal prompts. In this paper, we conduct a direct analysis of the multi-modal prompts by asking the following questions: $(i)$ How do the learned multi-modal prompts improve the recognition performance? $(ii)$ What do the multi-modal prompts learn? To answer these questions, we begin by isolating the component of the formula where the prompt influences the calculation of self-attention at each layer in two distinct ways, \ie, $(1)$ introducing prompt embeddings makes the $[cls]$ token focus on foreground objects. $(2)$ the prompts learn a bias term during the update of token embeddings, allowing the model to adapt to the target domain. Subsequently, we conduct extensive visualization and statistical experiments on the eleven diverse downstream recognition datasets. From the experiments, we reveal that the learned prompts improve the performance mainly through the second way, which acts as the dataset bias to improve the recognition performance of the pre-trained model on the corresponding dataset. Meanwhile, we propose the bias tuning way to validate our finding. With a deeper understanding of the multi-modal prompt, we hope our work can inspire new and solid research in this direction.
\end{abstract}    
\section{Introduction}
\label{sec:intro}
\begin{figure}[htbp]
    \centering
    \setlength{\abovecaptionskip}{5pt}
    \includegraphics[width=\linewidth]{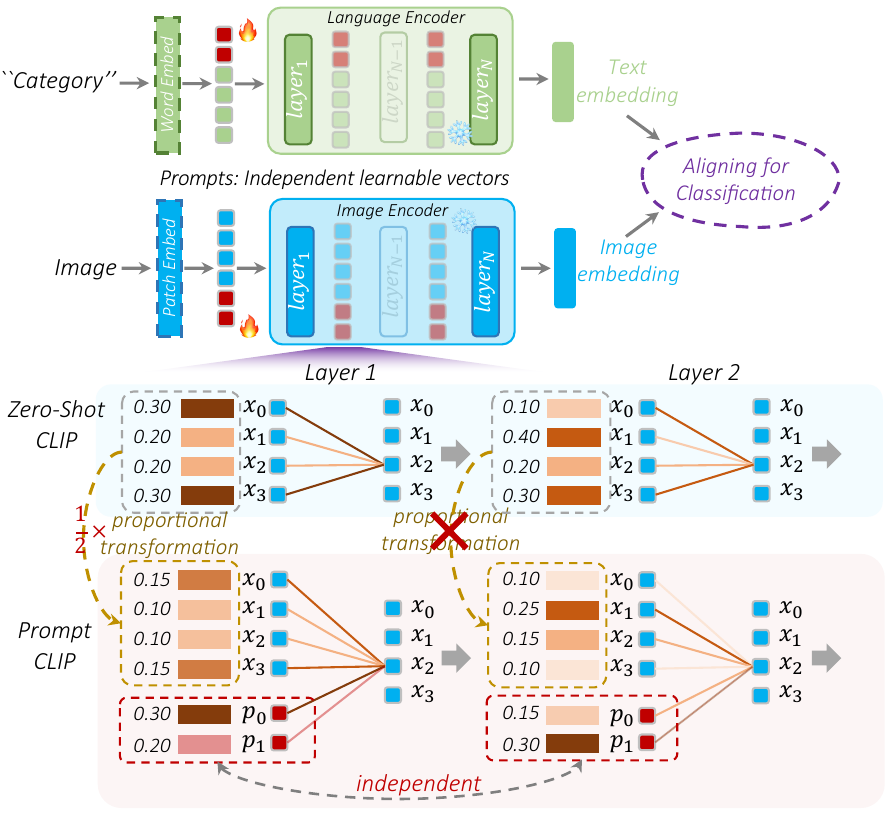}
    \caption{\textbf{Overscheme of the multi-modal prompts in the pre-trained model.} In each layer, the fixed prompts provide additional attention value for each token. In the first layer, the attention between tokens undergoes a proportional transformation (\red{$\frac{1}{2}\times$}). However, the relative values of attention between tokens are changed in the subsequent layers. Detailed explanation in Sec.\ref{3.2.1}.}
    \label{fig1}
    \vspace{-0.5cm}
\end{figure}

The pre-trained Vision-Language (VL) models \cite{radford2021learning, jia2021scaling, zhai2022lit, yao2021filip, yuan2021florence}  are pre-trained on a large corpus of image-text pairs available on the internet, such as CLIP400M and ALIGN1B, in a self-supervised manner.  They possess a strong understanding of open-vocabulary concepts, making them suitable for various downstream vision and vision-language applications \cite{gao2021clip, zhang2021tip, rasheed2022bridging, Maaz2022Multimodal, zhou2022detecting, gu2021open, manzoor2023multimodality, zang2022open, li2022language, rao2022denseclip, ding2022decoupling}. However, their zero-shot performance is unsatisfactory in many specific datasets, especially those with significant differences from natural language and images. It is crucial to tune these pre-trained models to specific datasets.

Prompt tuning \cite{zhou2022conditional,zhou2022learning, chen2022prompt, huang2022unsupervised, shu2022test, lu2022prompt,bahng2022visual,khattak2023maple,rasheed2023fine,khattak2023self} has emerged as a more efficient alternative to fine-tuning large-scale models. This approach introduces a small number of new learnable embeddings at the input, known as prompt tokens, to adapt the pre-trained models for downstream tasks while keeping the pre-trained model weights fixed, as shown in Fig.\ref{fig1}. Due to its efficiency in terms of parameters and convergence rate, prompt learning is found to be of great interest for adapting foundational models like CLIP for vision~\cite{jia2022visual,zhang2022neural,wang2022dualprompt,wang2022learning} and vision-language tasks~\cite{zhou2022learning,zhou2022conditional, zhu2022prompt, derakhshani2022variational}.  However, despite the progress in the development of prompt tuning, their design is still driven empirically, and a good understanding of their essential attribute is lacking.

This paper attempts to explain the mechanism of the learned multi-modal prompts from the perspective of attention statistics and visualization. We begin with leveraging the formula to separate the part of the prompts workspace in the encoder. We observe that prompts primarily work for the self-attention block of each layer in two ways, as shown in Fig.\ref{fig1}. $(i)$ prompts adapt the attention weights between the $[cls]$ token and the input tokens by assigning different attention weights to input tokens according to the similarity between the prompts and the corresponding tokens (detail in Sec.\ref{3.2.1}). $(ii)$ the learned prompts serve as the additional fixed tokens to provide fixed features to the $[cls]$ and input tokens.  We then probe the working ways of the prompts through the alignment contribution statistic and the separate analysis for image and language branches. 

In the alignment contribution statistic, we count the relevance between inputs and $[cls]$ token during alignment. In the statistics results of the text branch, we observe that the contribution of \texttt{category} tokens does not increase but decreases. The contribution of $t_{SOS}$ also decrease. On the contrary, the contribution of prompt tokens significantly increased. For the image branch, we reveal that the learned multi-modal prompts do not significantly adjust the relative relevance to $[cls]$ between inputs. Surprisingly, the contribution of multi-modal prompts to the alignment process is significant.

In the language branch, we find CLIP's textual branch primarily focuses on the $\bm{t}_{SOS}$ token, which serves as the bias. We attribute this as bias because the $\bm{t}_{SOS}$ token can not acquire the information from the input language features due to the casual self-attention mask mechanism. The learned text prompts enhance the model’s recognition performance on the relevant dataset by learning dataset biases and increasing the contribution of biases to the alignment task. 

In the vision branch, by probing the average attention distance, attention entropy and visualizing the attention of the $[cls]$ token, we observe that the prompts do not significantly alter the feature extraction process of the pre-trained model, which further supports the second working mechanism of prompts. Meanwhile, by examining the similarity heatmap of learned vision prompts, we find that the vision prompts resemble background features to the attention of the $[cls]$ token, \ie, the features that the $[cls]$ token did not pay attention to. Combining the experimental phenomena in contribution statistics, we find that the prompts can act as a \texttt{bridge} to enable the $[cls]$ token to focus on \texttt{background} features that it previously overlooked. Because the learned prompts are independent and not updated by the input features, we believe that visual prompts, similar to text prompts, also act as dataset bias, which is important to the alignment.

To validate our findings, we propose a novel bias tuning that directly incorporates the learnable bias to each transformer block to validate the importance of the dataset bias. Comparative experiments demonstrate that bias tuning outperforms prompt tuning with the same number of parameters.

\noindent\textbf{Main Findings and Contributions:} (1) We probe to understand the working mechanism of the multi-modal prompts for the pre-trained vision-language model, conducting a series of extensive experiments from various perspectives, including alignment relevance, attention, attention distance, attention entropy, and visualization. Our findings show that the prompts mainly learn as the dataset bias. (2) We propose a novel bias tuning that directly incorporates the learnable bias to each transformer block to validate our finding. 

\section{Related Works}

\subsection{Vision Language models} 
Foundational vision-language (VL) models \cite{radford2021learning, jia2021scaling, zhai2022lit, yao2021filip, yuan2021florence} leverage both visual and textual modalities to encode rich multi-modal representations. These models are pre-trained on a large corpus of image-text pairs available on the internet in a self-supervised manner. For instance, CLIP \cite{radford2021learning} and ALIGN \cite{jia2021scaling} utilize around 400M and 1B image-text pairs, respectively, to train their multi-modal networks. During pre-training, contrastive loss is commonly used as a self-supervision loss. This loss pulls together the features of paired images and texts while pushing away the unpaired image-text features. VL models possess a strong understanding of open-vocabulary concepts, making them suitable for various downstream vision and vision-language applications \cite{gao2021clip, zhang2021tip, rasheed2022bridging, Maaz2022Multimodal, zhou2022detecting, gu2021open, manzoor2023multimodality, zang2022open, li2022language, rao2022denseclip, ding2022decoupling}.  However, transferring these foundational models for downstream tasks without compromising on their original generalization ability still remains a major challenge. Our work aims to address this problem by proposing a novel regularization framework to adapt VL models via prompt learning.

\subsection{Prompt learning}
Prompt learning is an alternative fine-tuning method for transferring a model towards downstream tasks without re-learning the trained model parameters. This approach adapts a pre-trained model by adding a small number of new learnable embeddings at the input known as prompt tokens. Due to its efficiency in terms of parameters and convergence rate, prompt learning is found to be of great interest for adapting foundational models like CLIP for vision~\cite{jia2022visual,zhang2022neural,wang2022dualprompt,wang2022learning} and vision-language tasks~\cite{zhou2022learning,zhou2022conditional, zhu2022prompt, derakhshani2022variational}. CoOp \cite{zhou2022learning} fine-tunes CLIP by optimizing a continuous set of prompt vectors in its language branch for few-shot image recognition. Bahng \etal \cite{bahng2022visual} perform visual prompt tuning on CLIP by learning prompts on the vision branch. \cite{chen2022prompt} and \cite{lu2022prompt} propose to learn multiple sets of prompts for learning different contextual representations. CoCoOp \cite{zhou2022conditional} highlights the overfitting problem of CoOp and proposes to condition prompts based on visual features for improved performance on generalization tasks. MaPLe \cite{khattak2023maple} proposes a multi-modal prompt learning approach by learning hierarchical prompts jointly at the vision and language branches of CLIP for better transfer. Our approach builds on the naturally independent prompt tuning way \cite{rasheed2023fine} where prompts are learned at both the vision and language encoder of CLIP.

\section{Preliminaries}
\label{sec:background}
\subsection{Vision-Language Pre-trained Model (VLM)}\par
\begin{figure*}[htbp]
    \centering
    \includegraphics[width=\textwidth]{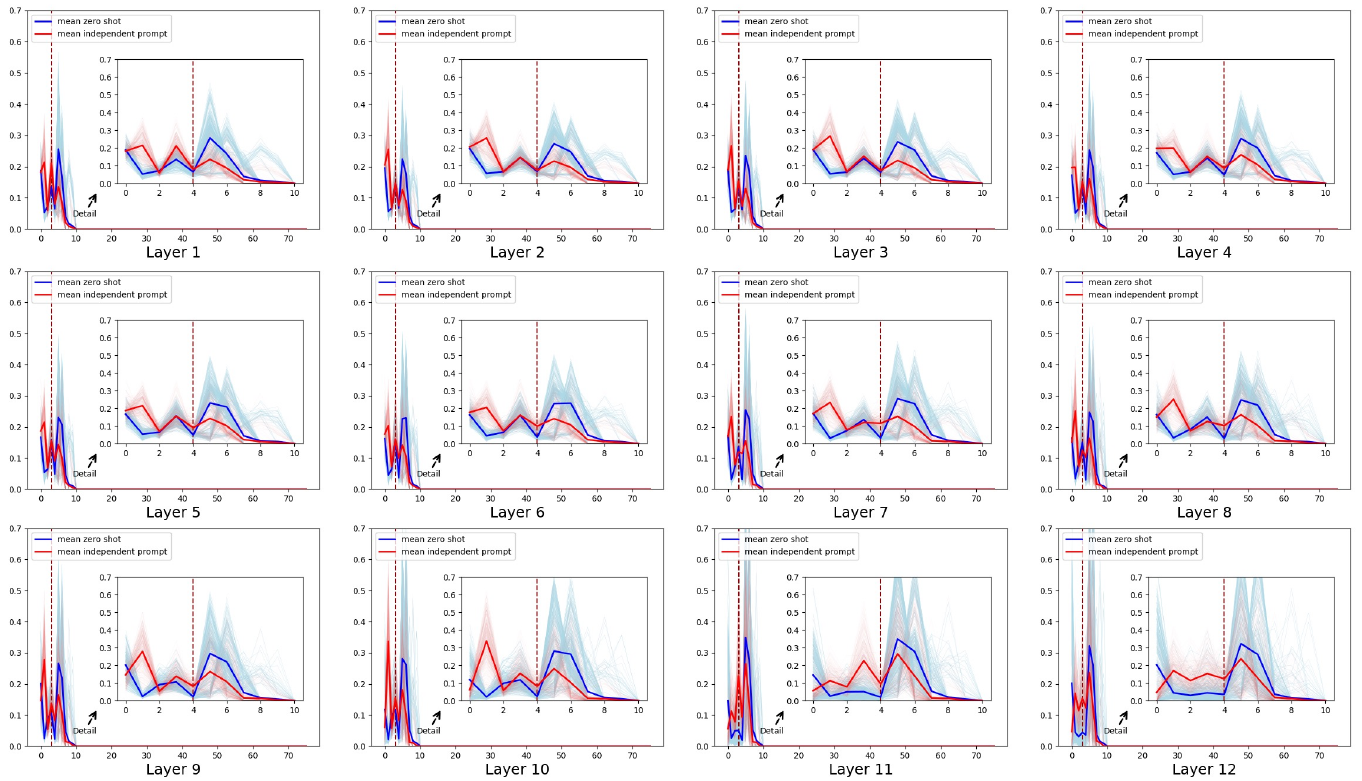}
    \caption{\textbf{Statistics results on the contribution of each token of the language branch to the alignment task for the OxfordPet dataset.} We conduct the statistical analysis on samples where the zero-shot CLIP misidentify, while the independent prompt CLIP identify correctly. The corresponding two realizations are averaged over the statistics for the entire dataset, where the \textbf{\blue{blue}} and \textbf{\red{red}} regions represent the statistics for zero-shot and prompt tuning, respectively. On the left side of the red vertical line, we show the contribution of $t_{SOS}$ and four learned text prompts $\bm{P}_{t}$. On the right side, we show the $\bm{c}_{k}$ tokens and padding tokens. The horizontal coordinate is the index of the token and the vertical coordinate is the relative value of the contribution. For the zero-shot clip, we use the temple: \texttt{``a photo of [Category].''}. For the tokens after the $\bm{t}_{EOS}$, we set their contribution as 0 for intuitive comparison. Meanwhile, we zoom in on comparing the index at which the prompt is placed in the middle of each figure. The statistics results of other datasets are detailedly shown in the Appendix.}
    \label{fig:oxford_pet_text}
\end{figure*}

\begin{figure*}[htbp]
    \centering
    \includegraphics[width=\textwidth]{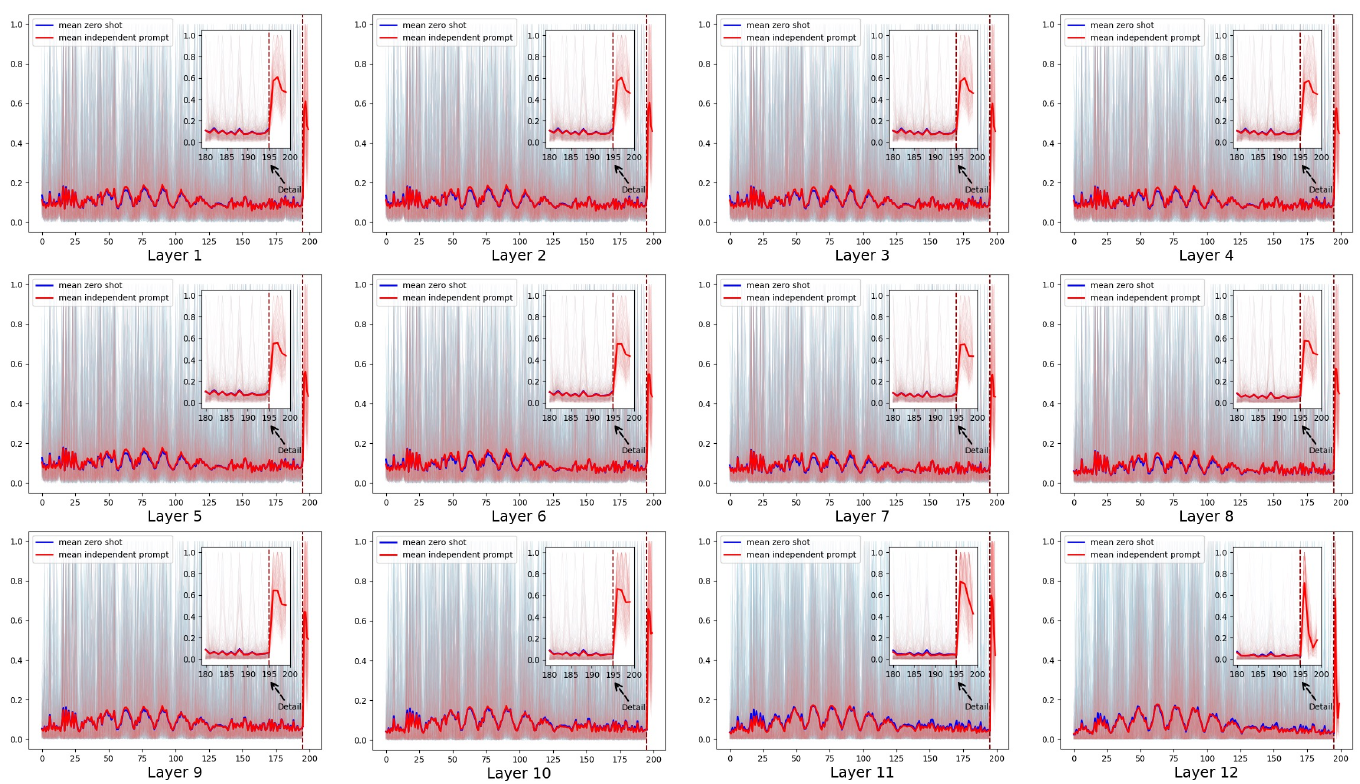}
    \caption{\textbf{Statistics results on the contribution of each patch of the image branch to the alignment task for the OxfordPet dataset.}  We conduct the statistical analysis on samples where the zero-shot CLIP misidentify, while the independent prompt CLIP identify correctly. The corresponding two realizations are averaged over the statistics for the entire dataset, where the \textbf{\blue{blue}} and \textbf{\red{red}} regions represent the statistics for zero-shot and prompt tuning, respectively. On the left side of the red vertical line, we show the contribution of the input patch tokens $\bm{\Tilde{X}_p}$. On the right side, we show the learned vision prompts $\bm{P}_{v}$. The horizontal coordinate is the index of the token and the vertical coordinate is the relative value of the contribution. For the zero-shot clip, the values of the 196th-199th index are empty due to the absence of the vision prompts. Meanwhile, we zoom in on comparing the index at which the prompt is placed in the middle of each figure. The statistics results of other datasets are detailedly shown in the Appendix.}
    \label{fig:oxford_pet_image}
\end{figure*}

We represent the vision and text encoders as ${f}$ and ${g}$, respectively. The pre-trained parameters are denoted as ${\theta}_{\mathtt{VLM}} = \{\theta_{f}, \theta_{g} \}$. The input image $\bm{X} \in \mathbb{R}^{C\times H\times W}$ is divided into $M$ patches $\left\{{patch}^1; \cdots ; {patch}^M\right\}$ followed by a projection $ \mathbf{E}\in \mathbb{R}^{\left(\frac{HW}{M} \cdot C\right) \times D_v}$ to produce patch tokens $\{\bm{x}_{1}, \bm{x}_{2}, \cdots, \bm{x}_{M}\}$.  Further, a learnable class token $\bm{x}_{cls}$ is attached to the input patches as $\bm{\Tilde{X}}= \{\bm{x}_{cls}, \bm{x}_{1}, \bm{x}_{2}, \cdots, \bm{x}_{M}\}$.  After adding the position embeddings $\bm{E}_{p o s} \in \mathbb{R}^{(M+1) \times D_v}$, the input patches are encoded by the encoder ${f}$ via $L_v$ transformer blocks in the bi-directional self-attention mask to produce the visual feature representation $\bm{\Tilde{f}} = f(\bm{\Tilde{X}}, \theta_{f})$, where $\bm{\Tilde{f}} \in  \mathbb{R}^{d}$. The detailed formulation is as follows: 
\begin{align}
\mathbf{x}_0 &=\left[\bm{x}_{cls} ; \bm{x}_{1}; \bm{x}_{2}; \cdots, \bm{x}_{M}\right]+\bm{E}_{p o s},\\
\mathbf{x}_{\ell}^{\prime} & =\mathtt{MSA}\left(\mathtt{LN}\left(\mathbf{x}_{\ell-1}\right)\right)+\mathbf{x}_{\ell-1}, \ell =1 \ldots L_v,\\
\mathbf{x}_{\ell}&=\mathtt{MLP}\left(\mathtt{LN}\left(\mathbf{x}_{\ell}^{\prime}\right)\right)+\mathbf{x}_{\ell}^{\prime}, \ell=1 \ldots L_v, \\ 
\bm{\Tilde{f}}&=\mathtt{LN}\left(\mathbf{x}_L^0\right) @ \ \mathbf{E_v}, \mathbf{E_v} \in \mathbb{R}^{{D_v}\times{d}}.
\end{align}
For the language branch, the corresponding category label ${y}$ is wrapped within a text template such as \texttt{‘a photo of a \{category\}’} which can be formulated as $\bm{\Tilde{Y}}=\{\bm{t}_{SOS}, \bm{t}_{1}, \bm{t}_{2}, \cdots, \bm{t}_{n}, \bm{c}_{k}, \bm{t}_{EOS}\}$. Here $\{\bm{t}_l|_{l=1}^{n}\}$ and $\bm{c}_{k}$ denote the word embeddings corresponding to the text template and the class label, respectively. $\bm{t}_{SOS}$  and  $\bm{t}_{EOS}$ are the learnable start and end token embeddings, respectively. Then, $\bm{\Tilde{Y}}$ are encoded by the text encoder ${g}$ via $L_t$ transformer blocks in the casual self-attention mask to the textual feature $\bm{\Tilde{g}} = g(\bm{\Tilde{Y}}, \theta_{g})$, where $\bm{\Tilde{g}} \in  \mathbb{R}^{d}$. The detailed formulation is as follows: 
\begin{align}
\mathbf{y}_0 &=\left[\bm{t}_{SOS}, \bm{t}_{1}, \bm{t}_{2}, \cdots, \bm{t}_{n}, \bm{c}_{k}, \bm{t}_{EOS}\right]+\bm{E}_{p o s},\\
\mathbf{y}_{\ell}^{\prime} & =\mathtt{MSA}\left(\mathtt{LN}\left(\mathbf{y}_{\ell-1}\right)\right)+\mathbf{y}_{\ell-1}, \ell =1 \ldots L_t,\\
\mathbf{y}_{\ell}&=\mathtt{MLP}\left(\mathtt{LN}\left(\mathbf{y}_{\ell}^{\prime}\right)\right)+\mathbf{y}_{\ell}^{\prime}, \ell=1 \ldots L_t, \\ 
\bm{\Tilde{g}}&=\mathbf{y}_L^{-1} @\ \mathbf{E_t},  \mathbf{E_t} \in \mathbb{R}^{{D_t}\times{d}}.
\end{align}
During the zero-shot inference, textual features of text template with class labels $\{c_1, c_2, \cdots, c_m\}$
are matched with image feature $\bm{\Tilde{f}}$, the category is predicted as follows:
\begin{align}
    \bm{pred} = \bm{argmax}(\frac{\mathtt{exp}(\mathtt{sim}(\bm{\Tilde{g}}, \bm{\Tilde{f}})\tau)}{\sum_{i=1}^{{num}_c}\mathtt{exp}(\mathtt{sim}(\bm{\Tilde{g^i}} \cdot \bm{\Tilde{f}})\tau)}),
\end{align}
 where $\mathtt{sim}(a,b)$ denotes the cosine similarity between $a$ and $b$, and $\tau$ is the temperature.

\subsection{Multi-Prompt Learning for VLM}\par
The multi-modal prompts learn hierarchical prompt tokens on both the text and image encoders separately. The independent multi-modal prompts append learnable $T$ language and $V$ visual prompts given as $\bm{P_{t}} = \{\bm{p_t}^1,\bm{p_t}^2, \cdots, \bm{p_t}^T\}$ and $\bm{P_{v}} = \{\bm{p_v}^1,\bm{p_v}^2, \cdots, \bm{p_v}^V\}$ with the textual categories and visual input tokens, respectively. The image encoder processes all tokens $\bm{\Tilde{X}_p}= \{\bm{e}_{cls}, \bm{e}_{1}, \bm{e}_{2}, \cdots, \bm{e}_{M}, \bm{p_v}^1,\bm{p_v}^2, \cdots, \bm{p_v}^V\}\}$ to generate prompted visual feature $\bm{\Tilde{f}_p}$. In the self-attention blocks, the prompts provide additional fixed features for the updating of the $[cls]$ token and patch tokens.

Similarly, the text branch formulation is omitted, and the textual feature $\bm{\Tilde{g}_p}$ is acquired, where $\bm{\Tilde{Y}_p}=\{\bm{t}_{SOS}, \bm{p^1_{t}},\bm{p}^{2}_{t}, \bm{p}^{3}_{t}, \cdots, \bm{p}^{T}_{t}, c_{k}, \bm{t}_{EOS}\}$. In this paper, we explore the independent deep multi-modal prompts.  The vision and language prompts are jointly represented as $ \bm{P} = \{ \bm{P_{v}}, \bm{P_{t}}\}$. For image classification on the downstream dataset $\mathcal{D}$, prompts $\bm{P}$ are inserted into the pre-trained and frozen $\theta_{f}$ and $\theta_{g}$ and are optimized with the cross-entropy loss, $\mathcal{L_{\text{CE}}}$, as follows:
\begin{align}
\label{eq:LCE}
    \mathcal{L_{\text{CE}}} = \text{arg}&\min_{\bm{P}}\mathbb{E}_{(\bm{X}, {y})\sim\mathcal{D}} \, \mathcal{L} (\text{sim}(\bm{\Tilde{f}_p},\bm{\Tilde{g}_p}), y).
\end{align}
In the inference phase, the model predicts with the learned multi-prompts as follows:
\begin{align}
    \bm{pred} = \bm{argmax}(\frac{\mathtt{exp}(\mathtt{sim}(\bm{\Tilde{g}_p}, \bm{\Tilde{f}_p})\tau)}{\sum_{i=1}^{C}\mathtt{exp}(\mathtt{sim}(\bm{\Tilde{g_p^i}} \cdot \bm{\Tilde{f}_p})\tau)}).
\end{align}

This paper aims to investigate the mechanism of the multi-modal prompts through the perspective of attention through statistical and visualization experiments.
\section{Exploring Experimrnts}

\subsection{Experiment Setting}
\textbf{Implementation Details:} We use the ViT-B/16 CLIP model in all experiments. Following the existing methods \cite{zhou2022conditional,khattak2023maple}, we uniformly train a few epochs, \ie, 10, using the $SGD$ optimizer, for preventing the model from overfitting to the few-shot data. We use the $cosine$ learning scheduler, and the learning rate is uniformly set as 0.0025. Meanwhile, we set the first epoch as the warmup epoch, and the warmup type and warmup constant learning rate are set as constant and 1e-5, respectively. In this paper, we conduct the exploring experiments in the 16-shot setting. The batch size is uniformly set as 32, and we did not search for specific suitable hyperparameters for individual datasets to improve the performance.\par
\noindent \textbf{Datasets:} We use 11 image recognition datasets. The datasets cover multiple recognition tasks including ImageNet~\cite{deng2009imagenet} and Caltech101~\cite{fei2004learning} that consist of generic object categories. OxfordPets~\cite{parkhi2012cats}, StanfordCars~\cite{krause20133d}, Flowers102~\cite{nilsback2008automated}, Food101~\cite{bossard2014food}, and FGVCAircraft~\cite{maji2013fine} have fine-grained categories. SUN397~\cite{xiao2010sun} and UCF101~\cite{soomro2012ucf101} are specific for scene and action recognition, respectively. DTD~\cite{cimpoi2014describing} has diverse texture categories and EuroSAT~\cite{helber2019eurosat} consists of satellite images.
\begin{figure*}[htbp]
    \centering
    \includegraphics[width=\textwidth]{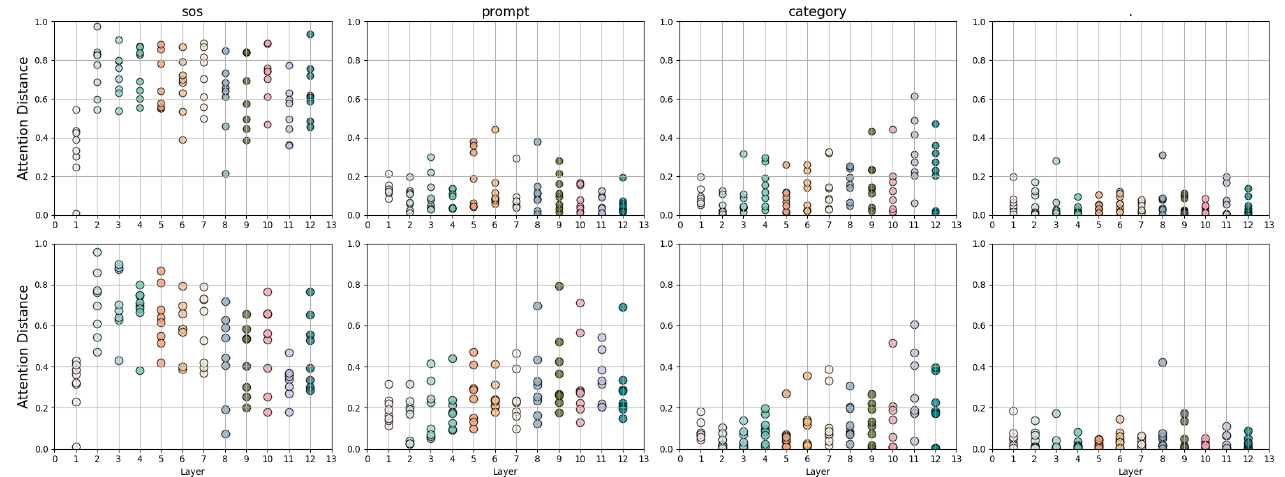}
    \caption{\textbf{Statistics results on the $[cls]$ token attention for the ImageNet dataset.} The first row is the attention distribution of CLIP, where the \texttt{prompt} denotes \texttt{`a photo of a'}. The second row is the feature distribution after adding the vision prompts, where the \texttt{prompt} represents the learnable token. Each graph contains the attention distribution of 12 transformer blocks, and each data layer includes 8 attention heads. \texttt{sos}, \texttt{prompt}, \texttt{category} and \texttt{`.'} denote the $[cls]$ token's attention on the corresponding token.} 
    \label{fig:text_attention_imagenet}
\end{figure*}

\subsection{How do the prompts improve the performance?}\label{3.2}
\subsubsection{Attention Formulation}\label{3.2.1}
Adding the multi-modal prompts, which learn from the few-shot training, often results in substantial recognition performance gains. Although the multi-modal prompts serve as tokens similar to the input tokens, it is also not rational to intuitively conclude how they work, since they are all at the high latitude abstraction space level, and we cannot determine what kind of image or text is learned. For the vision/text encoder, which consists of self-attention and MLP block, we analyze the impact of the added prompts in turn. For the MLP that attributes with independent operations for inputs, the added prompts through concatenate operations do not affect the upgrade of the input tokens. For self-attention layers, the prompts affect the attention weights of the $[cls]$ and input tokens in the attention calculation. We take the visual branch as an example to derive, after adding the prompt token  $\bm{\Tilde{X}}= \{\bm{x}_{cls}, \bm{x}_{1}, \bm{x}_{2}, \cdots, \bm{x}_{M}\}$ $\rightarrow$ $\bm{\Tilde{X}_p}= \{\bm{x}_{cls}, \bm{x}_{1}, \bm{x}_{2}, \cdots, \bm{x}_{M}, \bm{p_v}^1,\bm{p_v}^2, \cdots, \bm{p_v}^V\}\}$ (as shown in the Sec.\ref{sec:background}, in the image branch $\bm{x_0}$ denotes $\bm{x_{cls}}$). The updated process changes as follows: 
\begin{equation}
\begin{aligned}
\bm{x_i^{\ell+1}}&=\bm{x_i^{\ell}+\sum_{j=0}^M \underline{\underline{a_{i j}}} \cdot x_j}, \\
&\ \ \quad \quad \downarrow \\
\bm{x_i^{\ell+1}}&=\bm{x_i^{\ell}+\sum_{j=0}^M \underline{\underline{a_{i j}^{\prime}}} \cdot x_j}  \bm{+} \underline{\underline{\bm{\sum_{k=0}^V a_{i k}^{\prime} \cdot p_v^k}}},
\end{aligned}\label{12}
\end{equation}

Where $\bm{a_{ij}},\bm{a^{\prime}_{ij}} \propto \bm{e^{x_i A x_j}}$, $\bm{a^{\prime}_{ik}} \propto \bm{e^{x_i A p^k_v}}$ and $A=\frac{\bm{QK}^T}{\bm{\sqrt{d}}}$, $\bm{Q}$, $\bm{K} \in  \mathbb{R}^{{D_v}\times{D_v}}$ are projection matrix. The detailed formulation is shown in the Appendix. For the deep independent multi-modal prompts, the prompts are independently learned at each layer; once learned, they become a fixed piece of information for all data in the dataset, similar to the concept of bias. However, we observe that the information provided by the prompts is closely related to the patch tokens. That is, the more similar $x_i$ is to $p_c$, the larger the attention value obtained. Therefore, the information obtained by tokens from prompts in each layer differs. This results in changes to the values of the tokens in the subsequent layers, and the magnitude of these changes varies. As a result, the relative attention in the subsequent layers between tokens also changes. Thus, we make the following hypothesis about the working mechanism.
\begin{itemize}\setlength{\itemsep}{1pt}
\item[$\bullet$]Prompts improve the recognition performance mainly by adapting the attention values between the $[cls]$ token and the input tokens according to assigning different attention weights to input tokens. This means that the attention distribution of the $[cls]$ token towards the input tokens may be affected for better direction, leading to a better focus of the $[cls]$ token on foreground objects, which is crucial for classifying individuals.
\item[$\bullet$]Prompts mainly improve recognition performance by learning a bias term during the update of token embeddings, allowing the model to adapt to the target domain. This means fixed prompts may provide fixed dataset bias information at each layer after learning the corresponding dataset for better alignment between the image and language embeddings.
\item[$\bullet$]Both of the above are indispensable.
\end{itemize}
To investigate and validate the aforementioned hypothesis, we conducted a series of subsequent statistical and visualization experiments.
\begin{figure*}[t]
\centering
\begin{minipage}[htbp]{0.70\textwidth}
\includegraphics[width=1\linewidth]{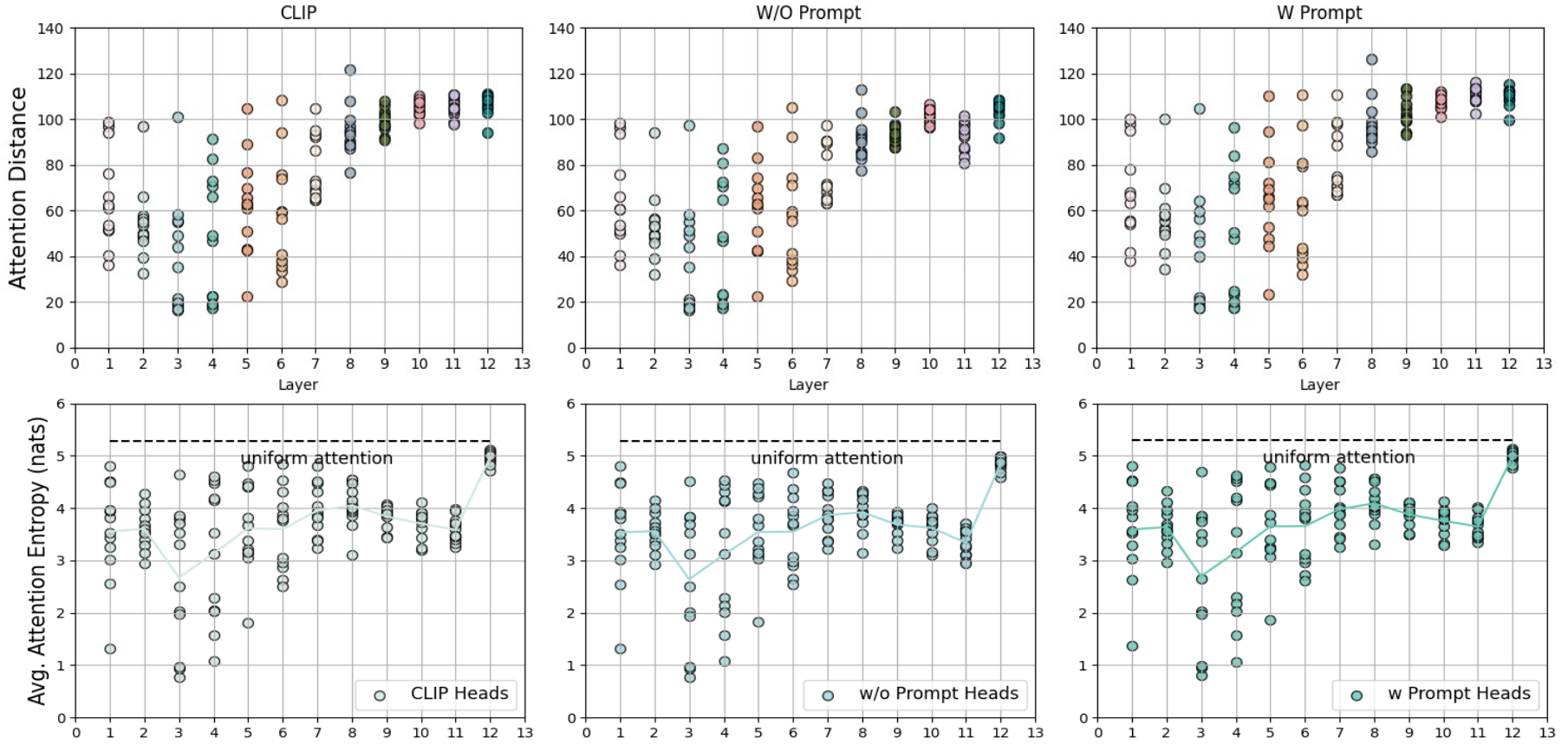}
\end{minipage}
\begin{minipage}[htbp]{0.28\textwidth}
\captionof{figure}{\textbf{The averaged attention distance and entropy in different attention heads (dots).} For the \texttt{CLIP}, we statistic the averaged attention distance and entropy of the original CLIP model. For the \texttt{W/O Prompt}, we statistic the averaged attention distance and entropy of the CLIP model after adding the prompts, but we remove the attention distance of the prompts. For the \texttt{W Prompt}, the attention distance and entropy of the prompts are retained.}
\label{fig:attention_distance}
\end{minipage}
\end{figure*}

\subsubsection{Aligning Contribution Statistics}\label{3.2.2}
We start by examining the contributions of each patch and token in the alignment process of the corresponding image-text pairs. For the contribution of each branch, we statistic the relevance of all tokens, including prompt tokens, to the $[cls]$ token before and after adding the learned prompts. Following \cite{vis1,vis2}, we construct a relevancy map per interaction, \ie $\mathtt{R}^{t t}$ and $\mathtt{R}^{v v}$ for language and vision modality, respectively. We calculate the relevancy maps by the forward pass on the attention layers, where each layer contributes to the aggregated relevance matrices. Before each modality self-attention operation, each patch/token is self-contained. Thus, the self-attention interactions are initialized with the identity matrix as follows:
\begin{equation}
\mathtt{R}_0^{v v}=\mathbb{I}^{v \times v}, \quad \mathtt{R}_0^{t t}=\mathbb{I}^{t \times t}.
\end{equation}

Immediately following each modality's self-attention blocks, we utilize attention maps $\mathtt{A}=\operatorname{softmax}\left(\frac{\mathtt{Q} \cdot \mathtt{K}^{\top}}{\sqrt{d_h}}\right)$ to update the relevance matrices. $\mathtt{A}$ intuitively defines connections between each pair of tokens (in language modality) / patches (in vision modality). The relevancies are accumulated by adding each layer's contribution to the aggregated relevancies due to the residual connection in the transformer blocks. All $h$ across head attention maps are averaged by the gradients that follow \cite{vis1,vis2}. Meanwhile, the simple average across heads would result in distorted relevancy maps because attention heads differ in importance and relevance. Thus, we follow \cite{vis2} and define the final attention map $\overline{\mathtt{A}}$ as follows:
\begin{equation}
\overline{\mathtt{A}}=\mathbb{E}_h\left((\nabla \mathtt{A} \odot \mathtt{A})^{+}\right),
\end{equation}
where $\mathbb{E}_h$ represents the mean across the head dimension, $\odot$ denotes the Hadamard product, $\nabla \mathtt{A}:=\frac{\partial y_t}{\partial \mathtt{A}}$ for $y_t$ which is the model's output for the alignment target we wish to visualize. After obtaining the attention maps $\overline{\mathtt{A}}$ corresponding to each self-attention module of each modality, the final relevancy scores $\mathtt{R}^{v v}, \mathtt{R}^{t t}$ are as follows:
\begin{align}
\mathtt{R}_{\ell}^{v v}&=\mathtt{R}_{\ell-1}^{v v}+\overline{\mathtt{A}}_{\ell-1} \cdot \mathtt{R}_{\ell-1}^{v v}, \\
\mathtt{R}_{\ell}^{t t}&=\mathtt{R}_{\ell-1}^{t t}+\overline{\mathtt{A}}_{\ell-1} \cdot \mathtt{R}_{\ell-1}^{t t}.
\end{align}
Finally, to retrieve per-token/patch relevancies for the alignment, we consider the row corresponding to the $[cls]$ token in the vision modality and the $\bm{t}_{EOS}$ in the language modality, which describes the connection between the $[cls]$ or $\bm{t}_{EOS}$ token and the other image patches/text tokens, \ie $\mathtt{R}_{\ell}^{v v}\left[\mathtt{:},\ \mathtt{0},\ \mathtt{1:}\right]$, $\mathtt{R}_{\ell}^{t t}\left[\mathtt{:},\ \mathtt{-}1,\ \mathtt{:-1}\right]$.

To visually illustrate the dependency of alignment tasks on image patches and text tokens after incorporating prompts, we conduct a statistical analysis on eleven fine-grained recognition datasets. Before visualization, we preprocess the contribution according to normalization for intuitive comparison.

The distribution of relevance for the OxfordPet dataset is presented in Fig.\ref{fig:oxford_pet_text} and Fig.\ref{fig:oxford_pet_image}. 
\begin{itemize}
    \item In the statistics results of the text branch, we observe that the contribution of \texttt{category} tokens does not increase but decreases. The contribution of $t_{SOS}$ decreases, too. On the contrary, the contribution of the independent prompt tokens significantly increase.
    \item For the image branch, we reveal that the learned multi-modal prompts do not significantly adjust the relative relevance to $[cls]$ between inputs. Although there may be differences in the absolute values between tokens or patches in the statistical table, this is mainly due to normalization operations, and the actual relative changes are minimal. On the contrary, the contribution of multi-modal prompts to the alignment process is unusually prominent.
\end{itemize}
Therefore, from the alignment contribution statistics, we observe that the mechanism of prompts better aligns with the second way in Sec.\ref{3.2.1}. Below, we further delve into the investigation by studying what prompts learn and how they influence the alignment process in the language and image branches, respectively.

\subsection{What do the multi-modal prompts learn?}
\begin{figure*}[h]
    \centering
    \setlength{\abovecaptionskip}{1pt}
    \includegraphics[width=\textwidth]{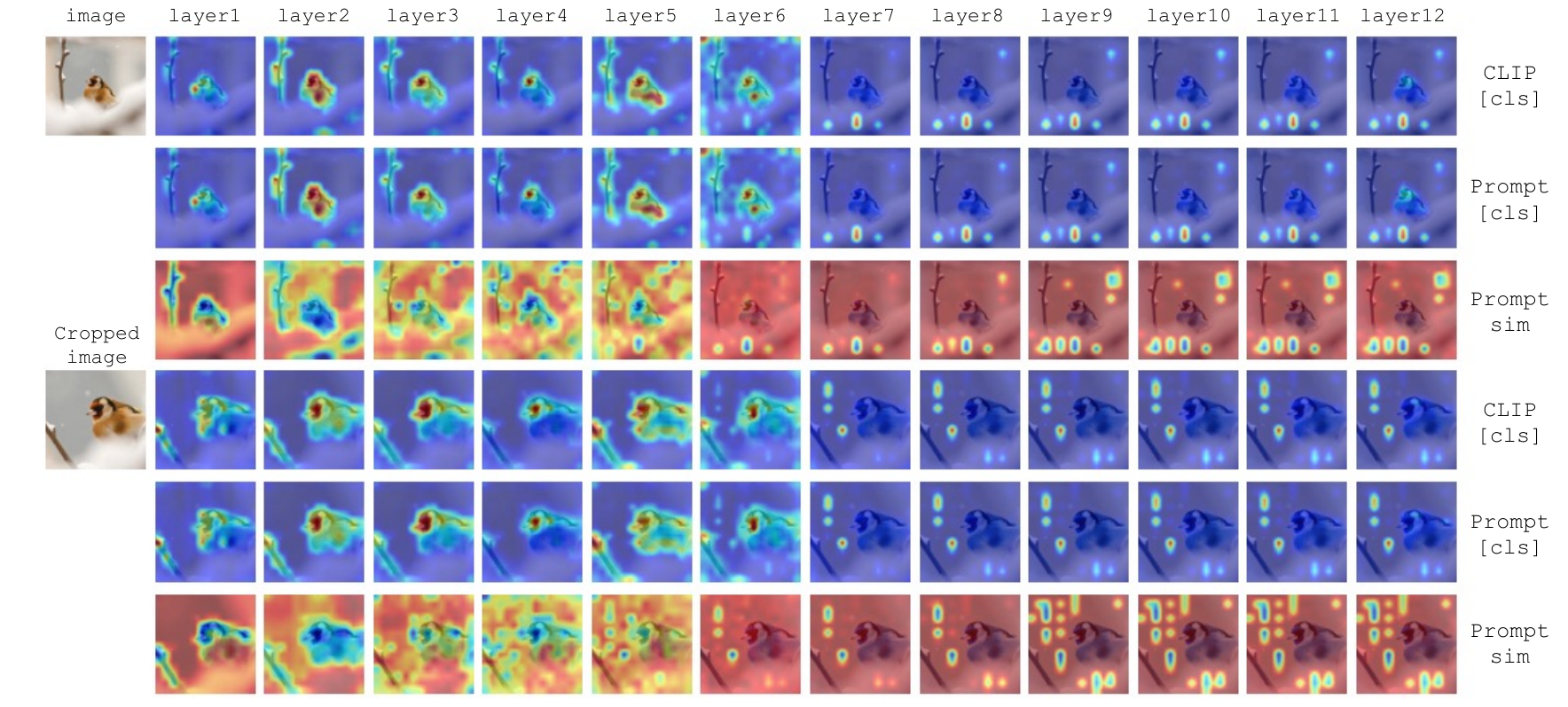}
    \caption{\textbf{Visualization results on the $[cls]$ token attention and vision prompts similarity on input patches for the ImageNet dataset.} Where the first and second row is the $[cls]$ token attention on the input patches of original CLIP and prompt CLIP, the layer corresponds to the encoder transformer block. The last row is the vision prompts' similarity with the input tokens.}
    \label{fig:visualization_imagenet}
\end{figure*}

The multi-modal prompts are situated within a high-dimensional space, assuming a role that is equivalent to that of input features (image patches/language tokens) of the same level. Hence, while it may not be possible to visualize the prompt as concrete images and text directly, we can showcase the prompt by visualizing the distribution of its similarity to the input features at the same level. In this section, we explore the text and vision prompts through attention and similarity.

\subsubsection{Learned Textual prompts on Language Branch} 
For the text encoder, CLIP leverages the attention way of GPT \cite{brown2020language}, where the former token can not focus on the latter tokens. Thus, the last token called $\bm{t}_{EOS}$ takes on the role of $[cls]$ token. In the language branch, we explore the textual prompts by the statistical attention distribution between the $\bm{t}_{EOS}$ takes and all other tokens, \ie, $\bm{t}_{SOS}$, $\bm{P}_{t}$ and $\bm{c}_{k}$, as shown in \ref{fig:text_attention_imagenet}. 

From the zero-shot CLIP attention statistics results in the first row of Fig.\ref{fig:text_attention_imagenet}, we observe that the $\bm{t}_{EOS}$ predominantly focuses most of its attention on the $\bm{t}_{SOS}$ token in each layer. However, the $\bm{t}_{SOS}$ token cannot attend to the subsequent category information, so its role is merely that of a bias. Meanwhile, We find that the text prompts play a role similar to the $\bm{t}_{SOS}$ token, as they also precede the category information and represent fixed information learned from the entire dataset without being updated based on the category information. The only difference is that the $\bm{t}_{SOS}$ token learns biases based on the entire language domain, as CLIP \cite{radford2021learning} is pre-trained on a massive amount of image-text pairs, while the text prompts only learn biases specific to the corresponding dataset.

We observe that after incorporating the learned text prompts, the attention of $\bm{t}_{EOS}$ towards $\bm{t}_{SOS}$ significantly decreases, while its attention towards the prompts noticeably increases. Meanwhile, the attention toward \texttt{category} tokens stays the same. Combining the findings from Sec.\ref{3.2.2} and the results in Fig.\ref{fig:oxford_pet_text}, we conclude that learned text prompts transfer some of the attention of $\bm{t}_{EOS}$ on $\bm{t}_{SOS}$ towards themselves, prompting the pre-trained model to shift some of the attention from the entire text domain biases to biases specific to the relevant dataset. As a result, it enhances the model's discriminative performance on the corresponding dataset.

\begin{wraptable}{R}{0.45\linewidth}
\setlength\tabcolsep{3pt}
\centering
\renewcommand{\arraystretch}{0.8}
\scriptsize
\vspace{-10pt}
\caption{\textbf{Validating the importance of bias.} We compare bias with prompt tuning with existing methods in the same training settings.}\label{validation}
\vspace{-7pt}
\label{tab:extension}
\resizebox{\linewidth}{!}{
\begin{tabular}{cccc} 
\toprule
\multirow{2}{*}{Method} & \multicolumn{3}{c}{Shot Setting} \\
                        & 4         & 8         & 16       \\ \midrule
VPT                     & 67.6      & 70.3      & 74.3     \\
CoOP                    & 69.4      & 72.9      & 76.6     \\
IVLP                    & 71.7      & 74.4      & 77.8     \\
MaPLe                   & 71.3      & 74.2      & 76.9     \\
TaskRes$\star$                 & 65.9      & 66.8      & 68.6     \\
Ours                    & 72.4      & 75.4      & 79.5    \\ \bottomrule
\end{tabular}}
\vspace{-15pt}
\end{wraptable}

\subsubsection{Vision prompts on Image Branch}
In the image branch, we explore the vision prompts in the following ways: (i) Probing the average attention distance. (ii) Probing the attention entropy of the attention weight and $[cls]$ attention. (iii) Visualizing $[cls]$ token's attention on the image patches and the similarity distribution between the vision prompts and image patches.

\textbf{Attention Distance:} To probe the impact of incorporating vision prompts on the feature extraction mechanism of the encoder, we analyze the average attention distance of the model before and after adding prompts, as shown in the first row of Fig.\ref{fig:attention_distance}. From the statistical results comparison of the \texttt{CLIP}, \texttt{W/O Prompt} and \texttt{W Prompt}, we observe that there is no significant impact on the model's attention distance before and after incorporating prompts for the input patches. This indicates that prompts do not significantly alter the feature extraction process of the pre-trained model. This further supports the second working mechanism of prompts in Sec.\ref{3.2.1}.

\textbf{Attention Entropy:} We analyze the model's attention before and after adding prompts and show the results in the second row of Fig.\ref{fig:attention_distance} for probing the impact of incorporating vision prompts on the vision encoder's feature extraction mechanism. The entropy value of attention reflects the focus of attention. A higher entropy value indicates a more uniform attention distribution and a global focus of attention. A lower entropy value indicates a more extreme attention distribution and a local focus of attention. From the results, we obtained the same conclusion as the attention distance statistic results: the prompts do not significantly impact the attention scope.

\textbf{visualization:} For the $[cls]$ token attention, we visualize the attention as the heatmap, as shown in Fig.\ref{fig:visualization_imagenet}. For the vision prompt similarity distribution, we calculate the Euclidean distance between the vision prompts and image patches. Due to the negative correlation between distance and similarity, after interpolation and normalization, we apply a linear transformation to the distance map by taking its complement, \ie, $\mathtt{x} = \mathtt{1} - \mathtt{x}$. To increase input diversity and showcase more detailed visual comparisons, we introduce rotation operations for the vision input, as shown in Fig.\ref{fig:visualization_imagenet}. From the visualization results, we observe that the vision prompts do not influence $[cls]$ token's attention on the input patches, which further supports the second working mechanism of prompts in Sec.\ref{3.2.1}.

Meanwhile, Fig.\ref{fig:visualization_imagenet} visually demonstrates that the learned vision prompts resemble background features in relation to the attention of the $[cls]$ token, \ie, the features that the $[cls]$ token did not pay attention to. Since Fig.\ref{fig:oxford_pet_image} shows that the attention of the $[cls]$ token towards the prompt is significant, we conclude that the prompt can act as a \texttt{bridge} to enable the $[cls]$ token to focus on \texttt{background} features that it previously overlooked. Furthermore, considering the fact that the learned prompts are independent and not updated by the input features, we believe that visual prompts, similar to text prompts, also act as dataset bias.

\subsection{Validation for the importance of the bias}

To validate the conclusion, we directly add the learnable bias to the pre-trained model and compare it with the prompt tuning. In the bias tuning setting, we add an independent learnable embedding and repeat it to add to all tokens after each transformer layer of the image and language branch. After learning in the same few-shot setting, we compare the bias tuning, prompt tuning, zero-shot CLIP, and some other existing methods in Table.\ref{validation}.

\section{Conclusion}
This paper probes to understand the working mechanism of the multi-modal prompts for the pre-trained vision-language model, conducting a series of extensive experiments from various perspectives, including alignment relevance, attention, attention distance, attention entropy, and visualization. Our findings show that the prompts mainly learn as the dataset bias. We propose a novel bias tuning that directly incorporates the learnable bias to each transformer block to validate our finding.
\section*{Acknowledgment}
This work is supported by the National Nature Science Foundation of China (grant No.61871106) and the Open Project Program Foundation of the Key Laboratory of Opto- Electronics Information Processing, Chinese Academy of Sciences (OEIP-O-202002).\\

{
    \small
    \bibliographystyle{ieeenat_fullname}
    \bibliography{main}
}

\end{document}